\documentclass[11pt,a4paper]{article}
\usepackage[hyperref]{naaclhlt2019}
\usepackage{times}
\usepackage{latexsym}

\usepackage{amsmath}
\usepackage{multirow}
\usepackage{url}

\usepackage{graphicx}
\usepackage{amsfonts}
\usepackage{bm}
\usepackage{microtype}

\aclfinalcopy 
\def\aclpaperid{2199} 

\title{Positional Encoding to Control Output Sequence Length}

\author{Sho Takase \hspace{1.5em} Naoaki Okazaki \\
  Tokyo Institute of Technology \\
  {\tt \{sho.takase@nlp.c, okazaki@c\}.titech.ac.jp} \\
  }

\date{}

\begin{document}
\maketitle
\begin{abstract}
Neural encoder-decoder models have been successful in natural language generation tasks.
However, real applications of abstractive summarization must consider additional constraint that a generated summary should not exceed a desired length.
In this paper, we propose a simple but effective extension of a sinusoidal positional encoding~\cite{NIPS2017_7181} to enable neural encoder-decoder model to preserves the length constraint.
Unlike in previous studies where that learn embeddings representing each length, the proposed method can generate a text of any length even if the target length is not present in training data.
The experimental results show that the proposed method can not only control the generation length but also improve the ROUGE scores.
\end{abstract}

\section{Introduction}
\begin{figure}[!t]
  \centering
  \includegraphics[width=7.5cm]{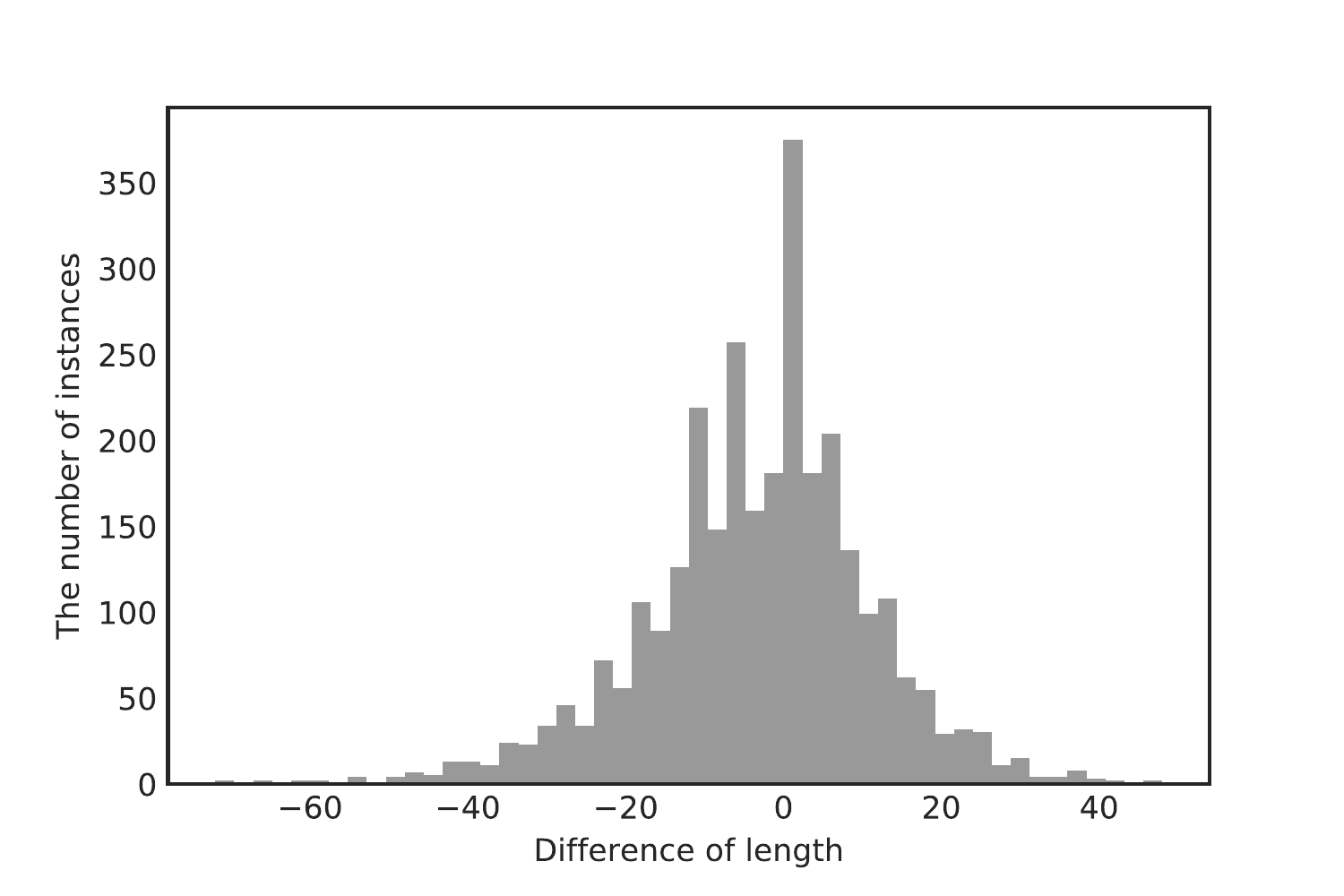}
   \caption{Difference in number of characters between correct headlines and outputs of a widely used LSTM encoder-decoder~\cite{luong-pham-manning:2015:EMNLP} which is trained on sentence-headline pairs created by \newcite{rush-chopra-weston:2015:EMNLP} from the annotated English Gigaword corpus. The difference was investigated for 3,000 sentence-headline pairs randomly sampled from the test splits.}
   \label{fig:length_dist}
\end{figure}

Neural encoder-decoder models have been successfully applied to various natural language generation tasks including machine translation~\cite{Sutskever:2014:SSL:2969033.2969173}, summarization~\cite{rush-chopra-weston:2015:EMNLP}, and caption generation~\cite{journals/corr/VinyalsTBE14}.
Still, it is necessary to control the output length for abstractive summarization, which generates a summary for a given text while satisfying a space constraint.
In fact, Figure \ref{fig:length_dist} shows a large variance in output sequences produced by a widely used encoder-decoder model~\cite{luong-pham-manning:2015:EMNLP}, which has no mechanism for controlling the length of the output sequences.

\newcite{W18-2706} trained embeddings that correspond to each output length to control the output sequence length.
Since the embeddings for different lengths are independent, it is hard to generate a sequence of the length that is infrequent in training data.
Thus, a method that can model any lengths continuously is required.

\newcite{kikuchi-EtAl:2016:EMNLP2016} proposed two learning based methods for an LSTM encoder-decoder: LenEmb and LenInit.
LenEmb inputs an embedding representing the remaining length in each decoding step.
Since this approach also prepares embeddings for each length independently, it suffers from the same problem as that in \newcite{W18-2706}.

On the other hand, LenInit can handle arbitrary lengths because it combines the scalar value of a desired length with a trainable embedding.
LenInit initializes the LSTM cell of the decoder with the embedding depending on the scalar value of the desired length.
\newcite{D18-1444} incorporated such scalar values into the initial state of the decoder in a CNN encoder-decoder.
These approaches deal with any length but it is reasonable to incorporate the distance to the desired terminal position into each decoding step such as in LenEmb.

In this study, we focused on Transformer~\cite{NIPS2017_7181}, which recently achieved the state-of-the-art score on the machine translation task.
We extend the sinusoidal positional encoding, which represents a position of each token in Transformer~\cite{NIPS2017_7181}, to represent a distance from a terminal position on the decoder side.
In this way, the proposed method considers the remaining length explicitly at each decoding step.
Moreover, the proposed method can handle any desired length regardless of its appearance in a training corpus because it uses the same continuous space for any length.

We conduct experiments on the headline generation task.
The experimental results show that our proposed method is able to not only control the output length but also improve the ROUGE scores from the baselines.
Our code and constructed test data are publicly available at: \href{https://github.com/takase/control-length}{https://github.com/takase/control-length}.

\section{Positional Encoding}
Transformer~\cite{NIPS2017_7181} uses a sinusoidal positional encoding to represent the position of an input.
Transformer feeds the sum of the positional encoding and token embedding to the input layer of its encoder and decoder.
Let $pos$ be the position and $d$ be the embedding size.
Then, the $i$-th dimension of the sinusoidal positional encoding $PE_{(pos, i)}$ is as follows:
\begin{align}
  PE_{(pos, 2i)} &= {\rm sin}\bigg(\frac{pos}{10000^{\frac{2i}{d}}}\bigg), \label{eq:sin}\\
  PE_{(pos, 2i+1)} &= {\rm cos}\bigg(\frac{pos}{10000^{\frac{2i}{d}}}\bigg). \label{eq:cos}
\end{align}
In short, each dimension of the positional encoding corresponds to a sinusoid whose period is $10000^{2i / d} \times 2\pi$.
Since this function returns an identical value at the same position $pos$, the above positional encoding can be interpreted as representing the absolute position of each input token.

In this paper, we extend Equations (\ref{eq:sin}) and (\ref{eq:cos}) to depend on the given output length and the distance from the terminal position.
We propose two extensions: length-difference positional encoding ($LDPE$) and length-ratio positional encoding ($LRPE$).
Then we replace Equations (\ref{eq:sin}) and (\ref{eq:cos}) with (\ref{eq:ctrl_diff_sin}) and (\ref{eq:ctrl_diff_cos}) (or (\ref{eq:ctrl_ratio_sin}) and (\ref{eq:ctrl_ratio_cos})) on the decoder side to control the output sequence length.
We define $LDPE$ and $LRPE$ as follows:
\begin{align}
  LDPE_{(pos, len, 2i)} &= {\rm sin}\bigg(\frac{len - pos}{10000^{\frac{2i}{d}}}\bigg), \label{eq:ctrl_diff_sin}\\
  LDPE_{(pos, len, 2i+1)} &= {\rm cos}\bigg(\frac{len - pos}{10000^{\frac{2i}{d}}}\bigg), \label{eq:ctrl_diff_cos} \\
  LRPE_{(pos, len, 2i)} &= {\rm sin}\bigg(\frac{pos}{len^{\frac{2i}{d}}}\bigg), \label{eq:ctrl_ratio_sin}\\
  LRPE_{(pos, len, 2i+1)} &= {\rm cos}\bigg(\frac{pos}{len^{\frac{2i}{d}}}\bigg), \label{eq:ctrl_ratio_cos}
\end{align}
where $len$ presents the given length constraint.
$LDPE$ returns an identical value at the position where the remaining length to the terminal position is the same.
$LRPE$ returns a similar value at the positions where the ratio of the remaining length to the terminal position is similar.
Let us consider the $d$-th dimension as the simplest example.
Since we obtain ${\rm sin}(pos / len)$ (or ${\rm cos}(pos / len)$) at this dimension, the equations yield the same value when the remaining length ratio is the same, e.g., $pos = 5$, $len = 10$ and $pos = 10$, $len = 20$.

We add $LDPE$ (or $LRPE$) to the input layer of Transformer in the same manner as in \newcite{NIPS2017_7181}.
In the training step, we assign the length of the correct output to $len$.
In the test phase, we control the output length by assigning the desired length to $len$.

\section{Experiments}
\subsection{Datasets}
We conduct experiments on the headline generation task on Japanese and English datasets.
The purpose of the experiments is to evaluate the ability of the proposed method to generate a summary of good quality within a specified length.
We used JAMUL corpus as the Japanese test set~\cite{Hitomi2019}.
This test set contains three kinds of headlines for 1,181\footnote{We obtained this test set by applying the pre-processing script at \href{https://github.com/asahi-research/Gingo}{https://github.com/asahi-research/Gingo} to the original JAMUL corpus.} news articles written by professional editors under the different upper bounds of headline lengths.
The upper bounds are 10, 13, and 26 characters ($len = 10, 13, 26$).
This test set is suitable for simulating the real process of news production because it is constructed by a Japanese media company.

In contrast, we have no English test sets that contain headlines of multiple lengths.
Thus, we randomly extracted 3,000 sentence-headline pairs that satisfy a length constraint from the test set constructed from annotated English Gigaword~\cite{napoles:2012:AG} by pre-processing scripts of \newcite{rush-chopra-weston:2015:EMNLP}\footnote{\href{https://github.com/facebookarchive/NAMAS}{https://github.com/facebookarchive/NAMAS}}.
We set three configurations for the number of characters as the length constraint: 0 to 30 characters ($len=30$), 30 to 50 characters ($len=50$), and 50 to 75 characters ($len=75$).
Moreover, we also evaluate the proposed method on the DUC-2004 task 1~\cite{Over:2007:DC:1284916.1285157} for comparison with published scores in previous studies.

\begin{table*}[!t]
  \centering
  \footnotesize
  \begin{tabular}{| l | r r r | r r r | r r r |} \hline
  & \multicolumn{3}{|c|}{$len = 10$} & \multicolumn{3}{c|}{$len = 13$} & \multicolumn{3}{c|}{$len = 26$} \\ \hline
  Model & \multicolumn{1}{c}{R-1} & \multicolumn{1}{c}{R-2} & \multicolumn{1}{c|}{R-L} & \multicolumn{1}{c}{R-1} & \multicolumn{1}{c}{R-2} & \multicolumn{1}{c|}{R-L} & \multicolumn{1}{c}{R-1} & \multicolumn{1}{c}{R-2} & \multicolumn{1}{c|}{R-L} \\ \hline
  \multicolumn{10}{|l|}{Baselines} \\ \hline
  LenInit & 38.08 & 17.72 & 36.84 & 41.83 & 19.53 & 39.22 & 47.07 & 22.02 & 38.36 \\
  LC & 35.88 & 15.73 & 34.80 & 40.28 & 18.86 & 38.16 & 42.62 & 19.38 & 35.61 \\
  Transformer & 34.63 & 15.48 & 33.02 & 43.94 & 21.35 & 40.77 & 46.43 & 23.03 & 38.10 \\ \hline
  \multicolumn{10}{|l|}{Proposed method} \\ \hline
  Transformer+$LDPE$ & 42.84 & 21.07 & 41.31 & 46.51 & 22.83 & 43.76 & 50.89 & 24.18 & 40.82 \\
  +$PE$ & 42.85 & 20.67 & 41.47 & 46.72 & 22.70 & 43.75 & {\bf 51.32} & {\bf 25.15} & {\bf 41.48} \\
  Transformer+$LRPE$ & 42.70 & 21.62 & 41.35 & {\bf 47.05} & {\bf 23.70} & {\bf 44.13} & 50.68 & 24.70 & 41.23 \\
  +$PE$ & {\bf 43.36} & {\bf 21.63} & {\bf 41.93} & 46.39 & 23.09 & 43.49 & 51.21 & 25.03 & 41.43 \\ \hline
  \multicolumn{10}{|l|}{Proposed method trained on the dataset without headlines consisting of target lengths} \\ \hline
  Transformer+$LDPE$ & 41.91 & 20.01 & 40.69 & 45.88 & 22.61 & 43.16 & 50.90 & 24.37 & 40.48 \\
  +$PE$ & 42.33 & 20.46 & 40.88 & 44.78 & 22.33 & 42.27 & 50.87 & 24.54 & 40.89 \\
  Transformer+$LRPE$ & 41.91 & 20.10 & 40.52 & 46.01 & 22.87 & 43.47 & 50.33 & 24.37 & 41.00 \\
  +$PE$ & 42.59 & 20.76 & 41.16 & 46.52 & 23.65 & 43.81 & 50.73 & 24.64 & 41.01 \\ \hline
  \end{tabular}
  \caption{Recall-oriented ROUGE scores for each length on Japanese test set. This test set contains three kinds of headlines, i.e., $len=10, 13, 26$, tied to a single article.\label{tab:jamul_result}}
\end{table*}

\begin{table*}[!t]
  \centering
  \footnotesize
  \begin{tabular}{| l | r r r | r r r | r r r |} \hline
  & \multicolumn{3}{|c|}{$len = 30$} & \multicolumn{3}{c|}{$len = 50$} & \multicolumn{3}{c|}{$len = 75$} \\ \hline
  Model & \multicolumn{1}{c}{R-1} & \multicolumn{1}{c}{R-2} & \multicolumn{1}{c|}{R-L} & \multicolumn{1}{c}{R-1} & \multicolumn{1}{c}{R-2} & \multicolumn{1}{c|}{R-L} & \multicolumn{1}{c}{R-1} & \multicolumn{1}{c}{R-2} & \multicolumn{1}{c|}{R-L} \\ \hline
  \multicolumn{10}{|l|}{Baselines} \\ \hline
  LenInit& 44.58 & 25.90 & 43.34 & 48.42 & 25.47 & 45.56 & 50.78 & 25.74 & 46.42 \\
  LC & 45.17 & 26.73 & 44.09 & 46.56 & 24.55 & 44.10 & 48.67 & 24.83 & 44.98 \\
  Transformer & 47.48 & {\bf 29.77} & 46.17 & 50.02 & {\bf 28.04} & 47.29 & 47.31 & 24.83 & 43.75 \\ \hline
  \multicolumn{10}{|l|}{Proposed method} \\ \hline
  Transformer+$LDPE$ & 47.26 & 26.98 & 45.77 & 50.21 & 26.13 & 47.15 & 53.99 & 27.78 & 49.24 \\
  +$PE$ & 48.13 & 27.18 & 46.43 & 50.29 & 25.97 & 47.17 & 53.65 & 27.65 & 49.06 \\
  Transformer+$LRPE$ & 48.79 & 28.77 & 47.17 & 50.09 & 26.08 & 46.91 & 53.91 & 27.82 & 49.15 \\
  +$PE$ & {\bf 49.23} & 29.26 & {\bf 47.68} & 50.41 & 26.37 & 47.39 & {\bf 54.21} & {\bf 27.84} & {\bf 49.38} \\ \hline
  \multicolumn{10}{|l|}{Proposed method trained on the dataset without headlines consisting of the target lengths} \\ \hline
  Transformer+$LDPE$ & 47.35 & 26.76 & 45.70 & 50.46 & 25.96 & 47.30 & 53.69 & 27.61 & 49.04 \\
  +$PE$ & 47.44 & 27.42 & 45.99 & 50.67 & 26.07 & 47.57 & 53.76 & 27.53 & 49.03 \\
  Transformer+$LRPE$ & 48.54 & 28.89 & 47.06 & 50.65 & 26.19 & 47.34 & 53.94 & 27.88 & 49.11 \\
  +$PE$ & 49.08 & 29.09 & 47.58 & {\bf 50.78} & 26.64 & {\bf 47.60} & 53.77 & 27.68 & 48.93 \\ \hline
  \end{tabular}
  \caption{Recall-oriented ROUGE scores for each length on test data extracted from annotated English Gigaword.\label{tab:engiga_result}}
\end{table*}

\begin{table}[!t]
  \centering
  \footnotesize
  \begin{tabular}{| l | r r r |} \hline
  Model & \multicolumn{1}{c}{R-1} & \multicolumn{1}{c}{R-2} & \multicolumn{1}{c|}{R-L} \\ \hline
  \multicolumn{4}{|l|}{Baselines} \\ \hline
  LenInit & 29.78 & 11.05 & 26.49 \\
  LC & 28.68 & 10.79 & 25.72 \\
  Transformer & 26.15 & 9.14 & 23.19 \\ \hline
  \multicolumn{4}{|l|}{Proposed method} \\ \hline
  Transformer+$LDPE$ & 30.95 & 10.53 & 26.79 \\
  +$PE$ & 31.00 & 10.78 & 27.02 \\
  +Re-ranking & 31.65 & 11.25 & 27.46 \\
  Transformer+$LRPE$ & 30.74 & 10.83 & 26.69 \\
  +$PE$ & 31.10 & 11.05 & 27.25 \\
  +Re-ranking & 32.29 & 11.49 & 28.03 \\
  +Ensemble (5 models) & {\bf 32.85} & {\bf 11.78} & {\bf 28.52} \\ \hline
  \multicolumn{4}{|l|}{Previous studies for controlling output length} \\ \hline
  \newcite{kikuchi-EtAl:2016:EMNLP2016} & 26.73 & 8.39 & 23.88 \\
  \newcite{W18-2706} & 30.00 & 10.27 & 26.43 \\ \hline
  \multicolumn{4}{|l|}{Other previous studies} \\ \hline
  \newcite{rush-chopra-weston:2015:EMNLP} & 28.18 & 8.49 & 23.81 \\
  \newcite{suzuki-nagata:2017:EACLshort} & 32.28 & 10.54 & 27.80 \\
  \newcite{zhou-EtAl:2017:Long} & 29.21 & 9.56 & 25.51 \\
  \newcite{li-EtAl:2017:EMNLP20174} & 31.79 & 10.75 & 27.48 \\
  \newcite{C18-1121} & 29.33 & 10.24 & 25.24 \\ \hline
  \end{tabular}
  \caption{Recall-oriented ROUGE scores in DUC-2004.\label{tab:duc_result}}
\end{table}

Unfortunately, we have no large supervision data with multiple headlines of different lengths associated with each news article in both languages.
Thus, we trained the proposed method on pairs with a one-to-one correspondences between the source articles and headlines.
In the training step, we regarded the length of the target headline as the desired length $len$.
For Japanese, we used the JNC corpus, which contains a pair of the lead three sentences of a news article and its headline~\cite{Hitomi2019}.
The training set contains about 1.6M pairs\footnote{We obtained this training set by applying the pre-processing script at \href{https://github.com/asahi-research/Gingo}{https://github.com/asahi-research/Gingo}.}.
For English, we used sentence-headline pairs extracted from the annotated English Gigaword with the same pre-processing script used in the construction of the test set.
The training set contains about 3.8M pairs.

In this paper, we used a character-level decoder to control the number of characters.
On the encoder side, we used subword units to construct the vocabulary~\cite{sennrich-haddow-birch:2016:P16-12,P18-1007}.
We set the hyper-parameter to fit the vocabulary size to about 8k for Japanese and 16k for English.

\subsection{Baselines}

\begin{table*}[!t]
  \centering
  \footnotesize
  \begin{tabular}{| l | r r r | r r r |} \hline
  & \multicolumn{6}{c|}{Variance} \\ \hline
  & \multicolumn{3}{c|}{Japanese dataset} & \multicolumn{3}{c|}{English Gigaword} \\ \hline
  Model & $len = 10$ & $len = 13$ & $len = 26$ & $len = 30$ & $len = 50$ & $len = 75$ \\ \hline
  \multicolumn{7}{|l|}{Baselines} \\ \hline
  LenInit & 0.047 & 0.144 & 0.058 & 0.114 & 0.112 & 0.091 \\
  LC & 0.021 & 0.028 & 0.040 & 0.445 & 0.521 & 0.871 \\
  Transformer & 181.261 & 115.431 & 38.169 & 193.119 & 138.566 & 620.887 \\ \hline
  \multicolumn{7}{|l|}{Proposed method} \\ \hline
  Transformer+$LDPE$ & {\bf 0.000} & {\bf 0.000} & {\bf 0.000} & {\bf 0.015} & 0.012 & 0.013 \\
  +$PE$ & 0.003 & 0.001 & 0.001 & 0.016 & {\bf 0.009} & {\bf 0.007} \\
  Transformer+$LRPE$ & 0.121 & 0.210 & 0.047 & 0.082 & 0.071 & 0.187 \\
  +$PE$ & 0.119 & 0.144 & 0.058 & 0.142 & 0.110 & 0.173 \\ \hline
  \multicolumn{7}{|l|}{Proposed method trained on the dataset without headlines consisting of the target lengths} \\ \hline
  Transformer+$LDPE$ & {\bf 0.000} & 0.002 & {\bf 0.000} & 0.018 & {\bf 0.009} & 0.009 \\
  +$PE$ & 0.021 & 0.001 & 0.003 & 0.021 & 0.013 & 0.010 \\
  Transformer+$LRPE$ & 0.191 & 0.362 & 0.043 & 0.120 & 0.058 & 0.133 \\
  +$PE$ & 0.183 & 0.406 & 0.052 & 0.138 & 0.081 & 0.154 \\ \hline
  \end{tabular}
  \caption{Variances of generated headlines.\label{tab:var_length}}
\end{table*}

We implemented two methods proposed by previous studies to control the output length and handle arbitrary lengths.
We employed them and Transformer as baselines.

\paragraph{LenInit}
\newcite{kikuchi-EtAl:2016:EMNLP2016} proposed LenInit, which controls the output length by initializing the LSTM cell $m$ of the decoder as follows:
\begin{align}
  m = len \times b,
\end{align}
where $b$ is a trainable vector.
We incorporated this method with a widely used LSTM encoder-decoder model~\cite{luong-pham-manning:2015:EMNLP}\footnote{We used an implementation at \href{https://github.com/mlpnlp/mlpnlp-nmt}{https://github.com/mlpnlp/mlpnlp-nmt}.}.
For a fair comparison, we set the same hyper-parameters as in \newcite{D18-1489} because they indicated that the LSTM encoder-decoder model trained with the hyper-parameters achieved a similar performance to the state-of-the-art on the headline generation.

\paragraph{Length Control (LC)}
\newcite{D18-1444} proposed a length control method that multiplies the desired length by input token embeddings.
We trained the model with their hyper-parameters.

\paragraph{Transformer}
Our proposed method is based on Transformer~\cite{NIPS2017_7181}\footnote{\href{https://github.com/pytorch/fairseq}{https://github.com/pytorch/fairseq}}.
We trained Transformer with the equal hyper-parameters as in the base model in \newcite{NIPS2017_7181}.

\subsection{Results}
Table \ref{tab:jamul_result} shows the recall-oriented ROUGE-1 (R-1), 2 (R-2), and L (R-L) scores of each method on the Japanese test set\footnote{To calculate ROUGE scores on the Japanese dataset, we used \href{https://github.com/asahi-research/Gingo}{https://github.com/asahi-research/Gingo}.}.
This table indicates that Transformer with the proposed method (Transformer+$LDPE$ and Transformer+$LRPE$) outperformed the baselines for all given constraints ($len=10, 13, 26$).
Transformer+$LRPE$ performed slightly better than Transformer+$LDPE$.
Moreover, we improved the performance by incorporating the standard sinusoidal positional encoding (+$PE$) on $len=10$ and $26$.
The results imply that the absolute position also helps to generate better headlines while controlling the output length.

Table \ref{tab:engiga_result} shows the recall-oriented ROUGE scores on the English Gigaword test set.
This table indicates that $LDPE$ and $LRPE$ significantly improved the performance on $len=75$.
Moreover, the absolute position ($PE$) also improved the performance in this test set.
In particular, $PE$ was very effective in the setting of very short headlines ($len=30$).
However, the proposed method slightly lowered ROUGE-2 scores from the bare Transformer on $len=30, 50$.
We infer that the bare Transformer can generate headlines whose lengths are close to 30 and 50 because the majority of the training set consists of headlines whose lengths are less than or equal to 50.
However, most of the generated headlines breached the length constraints, as explained in Section \ref{sec:analysis}.

To investigate whether the proposed method can generate good headlines for unseen lengths, we excluded headlines whose lengths are equal to the desired length ($len$) from the training data.
The lower parts of Table \ref{tab:jamul_result} and \ref{tab:engiga_result} show ROUGE scores of the proposed method trained on the modified training data.
These parts show that the proposed method achieved comparable scores to ones trained on whole training dataset.
These results indicate that the proposed method can generate high-quality headlines even if the length does not appear in the training data.

Table \ref{tab:duc_result} shows the recall-oriented ROUGE scores on the DUC-2004 test set.
Following the evaluation protocol~\cite{Over:2007:DC:1284916.1285157}, we truncated characters over 75 bytes.
The table indicates that $LDPE$ and $LRPE$ significantly improved the performance compared to the bare Transformer, and achieved better performance than the baselines except for R-2 of LenInit.
This table also shows the scores reported in the previous studies.
The proposed method outperformed the previous methods that control the output length and achieved the competitive score to the state-of-the-art scores.

Since the proposed method consists of a character-based decoder, it sometimes generated words unrelated to a source sentence.
Thus, we applied a simple re-ranking to each $n$-best headlines generated by the proposed method ($n=20$ in this experiment) based on the contained words.
Our re-ranking strategy selects a headline that contains source-side words the most.
Table \ref{tab:duc_result} shows that Transformer+$LRPE$+$PE$ with this re-ranking (+Re-ranking) achieved better scores than the state-of-the-art~\cite{suzuki-nagata:2017:EACLshort}.

\subsection{Analysis of Output Length}\label{sec:analysis}
Following \newcite{D18-1444}, we used the variance of the generated summary lengths against the desired lengths as an indicator of the preciseness of the output lengths.
We calculated variance ($var$) for $n$ generated summaries as follows\footnote{\newcite{D18-1444} multiplies Equation (\ref{eq:var}) by $0.001$.}:
\begin{align}
  var = \frac{1}{n} \sum_{i=1}^{n} |l_i - len|^{2}, \label{eq:var}
\end{align}
where $len$ is the desired length and $l_i$ is the length of the generated summary.

Table \ref{tab:var_length} shows the values of Equation (\ref{eq:var}) computed for each method and the desired lengths.
This table indicates that $LDPE$ could control the length of headlines precisely.
In particular, $LDPE$ could generate headlines with the identical length to the desired one in comparison with LenInit and LC.
$LRPE$ also generated headlines with a precise length but its variance is larger than those of previous studies in very short lengths, i.e., $len = 10$ and $13$ in Japanese.
However, we consider $LRPE$ is enough for real applications because the averaged difference between its output and the desired length is small, e.g., $0.1$ for $len = 10$.

The lower part of Table \ref{tab:var_length} shows the variances of the proposed method trained on the modified training data that does not contain headlines whose lengths are equal to the desired length, similar to the lower parts of Table \ref{tab:jamul_result} and \ref{tab:engiga_result}.
The variances for this part are comparable to the ones obtained when we trained the proposed method with whole training dataset.
This fact indicates that the proposed method can generate an output that satisfies the constraint of the desired length even if the training data does not contain instances of such a length.

\section{Conclusion}
In this paper, we proposed length-dependent positional encodings, $LDPE$ and $LRPE$, that can control the output sequence length in Transformer.
The experimental results demonstrate that the proposed method can generate a headline with the desired length even if the desired length is not present in the training data.
Moreover, the proposed method significantly improved the quality of headlines on the Japanese headline generation task while preserving the given length constraint.
For English, the proposed method also generated headlines with the desired length precisely and achieved the top ROUGE scores on the DUC-2004 test set.

\section*{Acknowledgments}
The research results have been achieved by ``Research and Development of Deep Learning Technology for Advanced Multilingual Speech Translation'', the Commissioned Research of National Institute of Information and Communications Technology (NICT), Japan.

\bibliography{naaclhlt2019.bbl}
\bibliographystyle{acl_natbib}

\end{document}